\documentclass{article}
\usepackage{spconf,amsmath,graphicx,comment}
\usepackage{amssymb,amsthm}
\usepackage{algorithm2e}
\usepackage{booktabs}
\usepackage{relsize}
\usepackage{lipsum}
\usepackage{tabularx}

\newcommand{\critic}{\text{critic}}
\newcommand{\RMSProp}{\text{RMSProp}}
\newcommand{\Optimizer}{\text{Optimizer}}

\newcommand{\clip}{\text{clip}}

\title{Robust Speech Recognition using Generative Adversarial Networks}
\name{Anuroop Sriram$^*$, Heewoo Jun$^*$, Yashesh Gaur, Sanjeev Satheesh
\thanks{$^*$ equal contribution.}
}
\address{Baidu Research, Sunnyvale, CA, USA}

\begin{document}

\maketitle
\begin{abstract}
This paper describes a general, scalable, end-to-end framework that uses the generative adversarial network (GAN) objective to enable robust speech recognition. Encoders trained with the proposed approach enjoy improved invariance by learning to map noisy audio to the same embedding space as that of clean audio. Unlike previous methods, the new framework does not rely on domain expertise or simplifying assumptions as are often needed in signal processing, and directly encourages robustness in a data-driven way. We show the new approach improves simulated far-field speech recognition of vanilla sequence-to-sequence models without specialized front-ends or preprocessing.
\end{abstract}
\begin{keywords}
automatic speech recognition, speech enhancement, generative adversarial networks
\end{keywords}
\section{Introduction}
\label{sec:intro}

Automatic speech recognition (ASR) is becoming increasingly more integral in our day-to-day lives enabling virtual assistants and smart speakers like Siri, Google Now, Cortana, Amazon Echo, Google Home, Apple HomePod, Microsoft Invoke, Baidu Duer and many more. While recent breakthroughs have tremendously improved ASR performance \cite{amodei2016ds2, xiong2016achieving} 
these models still suffer considerable degradation from reasonable variations in reverberations, ambient noise, accents and Lombard reflexes that humans have little or no issue recognizing.

Most of these problems can be mitigated by training the models on a large volume of data that exemplify these effects. However, in the case of non-stationary processes, such as accents, accurate data augmentation is most likely infeasible, and in general, collecting high quality datasets can be expensive and time-consuming. Past robust ASR literature has considered hand-engineered front-ends and data-driven approaches in an attempt to increase the value of relatively parsimonious data with desired effects \cite{zhang2017robust-asr,benzeghiba2007speech-variability}. While these techniques are quite effective in their respective operating regimes, they do not generalize well to other modalities in practice due to the aforementioned reasons. Namely, it is difficult to model anything beyond reverberation and background noise from the first principles. Existing techniques do not directly induce invariance for ASR or are not scalable.
And, due to the sequential nature of speech, alignments are needed to compare two different utterances of the same text.

In this work, we employ the generative adversarial network (GAN) framework \cite{goodfellow2014gan} to increase the robustness of seq-to-seq models \cite{bahdanau2015b} in a scalable, end-to-end fashion. The encoder component is treated as the generator of GAN and is trained to produce indistinguishable embeddings between noisy and clean audio samples. Because no restricting assumptions are made, this new robust training approach can in theory learn to induce robustness without alignment or complicated inference pipeline and even where augmentation is not possible. We also experiment with encoder distance objective to explicitly restrict the embedding space and demonstrate that achieving invariance at the hidden representation level is a promising direction for robust ASR.

The rest of the paper is organized as follows. Related work is documented in Section \ref{sec:related-work}. Section \ref{sec:robust-asr} defines our notations and details the robust ASR GAN. Section \ref{sec:setup} explains the experimental setup. Section \ref{sec:results} shows results on the Wall Street Journal (WSJ) dataset with simulated far-field effects. Finishing thoughts are found in Section \ref{sec:conclusion}.

\section{RELATED WORK}
\label{sec:related-work}

A vast majority of work in robust ASR deals with reverberations and ambient noise; \cite{zhang2017robust-asr} provides an extensive survey in this effort. One of the most effective approaches in this variability is to devise a strong front-end such as the weighted prediction error (WPE) speech dereverberation \cite{nakatani2010wpe,yoshioka2012wpe} and train the resulting neural network with realistic augmented data \cite{li2017ghomeam,kim2017ghomeff}.

A shift from more traditional signal processing techniques to more modern, data-driven methods was seen when the denoising autoencoder \cite{vincent2010sdae} was employed to induce invariance to reverberations \cite{
mimura2015dae}. This is novel in that the autoencoder is explicitly trained to predict the original audio features from a perturbed version convolved with an impulse response. While denoising autoencoder models for enhancing speech have been shown to improve perceptual quality of the produced speech, they have not demonstrated significant improvement for the task of speech recognition. This is because autoencoders are trained to reconstruct all aspects of the original audio, including many features that are not important for speech recognition, such as the voice and accent of the speaker, background noises etc. In fact, ASR systems learn to remove such artifacts of the input audio as they can hinder speech recognition performance. \cite{ravanelli2017nodnn} proposed multiple rounds of joint denoising and ASR training for each audio sample, but this approach is not scalable for large datasets.

A similar approach in spirit is to minimize the distance in the embedding space between clean and noisy audio. The intuition here is that the embedding distance is a measure of semantic similarity \cite{hadsell2006inv}. 
However, the perturbed speech may have a different time duration than the reference audio; dynamic time warping \cite{thiolliere2016dtw} can be used to approximate the alignment and compare sequences of varying lengths, but there is an increased computational overhead.

\cite{bousmalis2017using} uses the generative adversarial networks (GAN) for domain adaptation to make the simulated images look more realistic to improve the task of robotic hand grasping. GAN \cite{goodfellow2014gan} is an unsupervised learning framework, where the generator network learns to produce increasingly more realistic data in attempt to fool a competing discriminator. Because equilibrium is reached at a saddle point, it is notoriously hard to train. There have been many improvements to this technique. For example, Wasserstein GAN \cite{arjovsky2017wgan} uses the Earth-Mover distance to mitigate optimization issues. It is also less susceptible to architectural choices.

For speech, \cite{pascual2017segan} proposes a GAN based speech enhancement method called SEGAN but without the end goal of speech recognition. SEGAN operates on raw speech samples and hence it is computationally impractical for large scale experiments.


\section{ROBUST ASR}
\label{sec:robust-asr}

\subsection{Encoder distance enhancer}

As explained in Section \ref{sec:related-work}, denoising reconstruction and perceptual enhancement do not significantly improve ASR. A better approach would be to reconstruct only those aspects of the audio which are important for predicting the text spoken and ignore everything else. We hypothesize that the encoders of well trained ASR systems would learn to retain only this information from the input audio. Based on this idea, we propose a new sequence-to-sequence architecture for robust speech recognition that tries to match the output of the encoder for clean audio and noisy audio.

\begin{figure}[t!]
    \centering
    \includegraphics[width=0.46\textwidth]{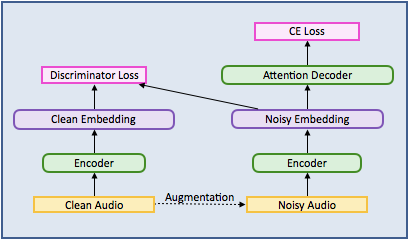}
    \caption{Architecture of the enhancer models introduced in this paper. The discriminator loss can be $L^1$-distance or WGAN loss. The entire model is trained end-to-end using both the discriminator loss and the cross-entropy loss. We use RIR convolution to simulate far-field audio. It's also possible to train this model with the same speech recorded in different conditions.}
    \label{fig:enhancer}
\end{figure}

The system works as follows: the same encoder, $g$, is applied to the clean audio $x$ and the corresponding noisy audio $\widetilde x$ to produce hidden states $z=g(x)$ and $\widetilde{z}=g(\widetilde x)$. The decoder, $h$, models the conditional probability $p(y|x) = p(y|z)$ 
and is used to predict the output text sequence one character at a time.
This architecture is described in Figure \ref{fig:enhancer}. 
The entire system is trained end-to-end using a multi-task objective that tries to minimize the cross-entropy loss of predicting $y$ from $\widetilde{x}$ and the normalized $L^1-$distance between $z$ and $\widetilde{z}$:
\begin{equation} \label{eq:dist_enh_loss}
\mathbb E_{(x,y) \sim \mathcal D} \left[
    H(h(\widetilde z), y) + \lambda \frac{\lVert z - \widetilde{z} \rVert_{1}}{\lVert z \rVert_{1} + \lVert \widetilde{z} \rVert_{1} + \epsilon }
\right].
\end{equation}

\subsection{GAN enhancer}

\begin{algorithm*}[htb!]
\KwData{$n_\critic$, the number of critic per robust ASR updates. $c$, the clipping parameter. $m$, the batch size.}
\While{\upshape $\theta$ has not converged}{
    \For{\upshape $t=1,\dots,n_\critic$} {
        Sample $\{(x^{(i)}, y^{(i)}) \sim \mathcal D\}_{i=1}^m$
        a batch of labeled speech data.
        \DontPrintSemicolon\;
        Sample $\{\widetilde x^{(i)}\}_{i=1}^m$
        by augmentation or from a different distribution.
        \DontPrintSemicolon\;
        Sample $\{\varepsilon^{(i)}\}_{i=1}^m$ a batch of prior noise.
        \DontPrintSemicolon\;
        $g_\theta \leftarrow \nabla_\theta \left[
            \frac{1}{m}\sum_{i=1}^m H(h_\theta(g_\theta(x^{(i)})), y^{(i)})
        \right]$
        \DontPrintSemicolon\;
        $\theta \leftarrow \theta - \Optimizer(\theta, g_\theta)$
        \DontPrintSemicolon\;
        $g_w \leftarrow \nabla_w \left[
            \frac{1}{m}\sum_{i=1}^m
            f_w(g_\theta(x^{(i)})) -
            \frac{1}{m}\sum_{i=1}^m
            f_w(g_\theta(\widetilde x^{(i)} + \varepsilon^{(i)}))
        \right]
        $
        \DontPrintSemicolon\;
        $w \leftarrow w + \RMSProp(w, g_w)$
        \DontPrintSemicolon\;
        $w \leftarrow \clip(w, -c, c)$
    }
    Sample $\{(x^{(i)}, y^{(i)}) \sim \mathcal D\}_{i=1}^m$
    a batch of labeled speech data.
    \DontPrintSemicolon\;
    Sample $\{\widetilde x^{(i)}\}_{i=1}^m$
    by augmentation or from a different distribution.
    \DontPrintSemicolon\;
    Sample $\{\varepsilon^{(i)}\}_{i=1}^m$ a batch of prior noise.
    \DontPrintSemicolon\;
    $g_\theta \leftarrow \nabla_\theta \left[
        \frac{1}{m}\sum_{i=1}^m H(h_\theta(g_\theta(x^{(i)})), y^{(i)})
        - \lambda \frac{1}{m}\sum_{i=1}^m
            f_w(g_\theta(\widetilde x^{(i)} +
                \varepsilon^{(i)}))
    \right]$
    \DontPrintSemicolon\;
    $\theta \leftarrow \theta - \Optimizer(\theta, g_\theta)$
}
\caption{WGAN enhancer training. The seq-to-seq model was trained using the Adam optimizer in our experiments. If $\widetilde x$ can be generated from $x$, data augmentation can also be used to update the seq-to-seq model.}
\label{alg:wgan-enhancer}
\end{algorithm*}

In our experiments, we found the encoder distance penalty to yield excellent results but it has the disadvantage that the encoder content between clean and noisy audio has to match frame for frame. Instead, employing the GAN framework, we can have a discriminator output a scalar likelihood of the entire speech being clean, and train the encoder to generate embeddings that are indistinguishable by the discriminator.

In this paper, Wasserstein GAN (WGAN) \cite{arjovsky2017wgan} is used. 
Following the notations of WGAN, we parametrize the seq-to-seq and discriminator models with $\theta$ and $w$ respectively. The overall architecture depicted in Figure \ref{fig:enhancer} remains the same, but the encoder distance in \eqref{eq:dist_enh_loss} is now replaced with the dual of Earth-Mover (EM) distance 
\begin{equation}
\max_{w\in \mathcal W} 
  \left\{
  \mathbb E_{x}
    \left[f_w(g_\theta(x))\right] -
  \mathbb E_{\widetilde x,\varepsilon}
    \left[f_w(g_\theta(\widetilde x + \varepsilon)\right]
  \right\}.
  \label{eq:em-loss}
\end{equation}
We treat the embedding of the clean input $x$ as real data and the embedding of $\widetilde x$, which can either be augmented from $x$ or drawn from a different modality, as being fake. And so, as GAN training progresses, the encoder $g_\theta$ should learn to remove extraneous information to ASR to be able to fool the discriminator. In practice, we found that including a random Gaussian noise $\varepsilon$ to the input prior of the generator helps improve training. Also, weights in the parameter set $\mathcal W$ should be clipped to ensure the duality of \eqref{eq:em-loss} holds up to a constant multiple \cite{arjovsky2017wgan}. The adapted WGAN training procedure is detailed in Algorithm \ref{alg:wgan-enhancer}.

\section{EXPERIMENTAL SETUP}
\label{sec:setup}

\subsection{Corpora and Tasks}
\label{sec:corpora-tasks}

We evaluated the enhancer framework on the Wall Street Journal (WSJ) corpus with simulated far-field effects. The dev93 and eval92 sets were used for hyperparameter selection and evaluation respectively. The reverberant speech is generated with room impulse response (RIR) augmentation as in \cite{ko2017reverberant}, where each audio is convolved with a randomly chosen RIR signal. The clean and far-field audio durations are kept the same with valid convolution so that the encoder distance enhancer can be applied. We collected 1088 impulse responses, using a linear array of 8 microphones, 120 and 192 of which were held out for development and evaluation. The speaker was placed in a variety of configurations, ranging from 1 to 3 meters distance and 60 to 120 degrees inclination with respect to the array, for 20 different rooms. Mel spectrograms of 20 ms samples with 10 ms stride and 40 bins were used as input features to all of our baseline and enhancer models.

\subsection{Network Architecture}
\label{sec:network-architecture}

\begin{table}[h!]
    \centering
    \begin{tabular}{c}
    \toprule
    Bidirectional GRU (dimension = 256, batch norm) \\
    Pooling (2x1 striding) \\
    Bidirectional GRU (dimension = 256, batch norm) \\
    Pooling (2x1 striding) \\
    Bidirectional GRU (dimension = 256, batch norm) \\
    Pooling (2x1 striding) \\
    Bidirectional GRU (dimension = 256, batch norm) \\
    Bidirectional GRU (dimension = 256, batch norm) \\
    Bidirectional GRU (dimension = 256, batch norm) \\
    \bottomrule
    \end{tabular}
    \caption{Architecture of the encoder.}
    \label{tab:encoder_arch}
\end{table}

For the acoustic model, we used the sequence-to-sequence framework with soft attention based on \cite{bahdanau2015b}. The architecture of the encoder is described in Table \ref{tab:encoder_arch}. The decoder consisted of a single 256 dimensional GRU layer with a hybrid attention mechanism similar to the models described in \cite{Battenberg2017ExploringNT}. 

\begin{table}[h!]
    \centering
    \begin{tabular}{c}
    \toprule
    7x2 Convolution, 32 filters, 5x1 striding \\
    3x3 Convolution, 64 filters, 2x1 striding \\
    Bidirectional LSTM (dimension = 32) \\
    3x3 Convolution, 64 filters, 2x1 striding \\
    3x3 Convolution, 96 filters, 1x1 striding \\
    Bidirectional LSTM (dimension = 32)\\
    Linear projection to per-time step scalar\\
    Sigmoid \\
    Mean pool of likelihood scores \\
    \bottomrule
    \end{tabular}
    \caption{Architecture of the critic. (feature)$\times$(time).}
    \label{tab:critic_arch}
\end{table}

The discriminator network of the WGAN enhancer is described in Table \ref{tab:critic_arch}. All convolutional layers use leaky ReLU activation \cite{maas2013leakyrelu} with 0.2 slope for the leak, and batch normalization \cite{ioffe2015bn}.

\begin{table*}[ht!]
    \centering
    \begin{tabular}{|l|cc|cc|}
        \toprule
         Model & \multicolumn{2}{c|}{Near-Field} & \multicolumn{2}{c|}{Far-Field} \\
         & CER & WER & CER & WER \\
        \midrule
        seq-to-seq & 7.43\% & 21.18\% & 23.76\% & 50.84\% \\
        seq-to-seq + far-field Augmentation & 7.69\% & 21.32\% & 12.47\% & 30.59\% \\
        seq-to-seq + $L^1$-Distance Penalty & 7.54\% & 20.45\% & 12.00\% & 29.19\% \\
        seq-to-seq + GAN Enhancer & 7.78\% & 21.07\% & \textbf{11.26\%} & \textbf{28.12\%} \\
        \bottomrule
         
    \end{tabular}
    \caption{Speech recognition performance on the Wall Street Journal Corpus}
    \label{tab:enhancer_results}
\end{table*}

\subsection{Training}
\label{sec:training}

To establish a baseline, in the first experiment, we trained a simple attention based seq-to-seq model. All the seq-to-seq networks in our experiments were trained using the Adam optimizer.
We evaluate all models on both clean and far-field test sets.

To study the effects of data augmentation, we train a new seq-to-seq model with the same architecture and training procedure as the baseline. However this time, in each epoch, we randomly select 40\% of the training utterances and apply the train RIRs to them (in our previous experiments we had observed that 40\% augmentation results in the best validation performance). 

For the enhancer models, $\lambda$ in Equation \ref{eq:dist_enh_loss} was tuned over the dev set by doing a logarithmic sweep in [0.01, 10]. $\lambda = 1$ gave the best performance.

We use Algorithm \ref{alg:wgan-enhancer} to train the WGAN enhancer. The clipping parameter was 0.05 and $\varepsilon$ was random normal with 0.001 standard deviation. We found that having a schedule for $n_\critic$ was crucial. Namely, we do not update the encoder parameters with WGAN gradients for the first 3000 steps. Then, we use the normal $n_\critic=5$. We hypothesize that the initial encoder embedding is of poor quality and encouraging invariance at this stage through the critic gradients significantly hinders seq-to-seq training.

\section{RESULTS}
\label{sec:results}
We present results in Table \ref{tab:enhancer_results}. All of the evaluations were performed using greedy decoding and no language models. To provide context, our near-field result is comparable to the 18.6\% WER of \cite{bahdanau2015b} obtained with language model beam decoding with 200 beam size.  We can see that a seq-to-seq model trained only on near-field audio data performs extremely poorly on far-field audio. This suggests that it is non-trivial for an ASR model to generalize from homogeneous near-field audio to far-field audio.

To overcome this, we train a stronger baseline with simulated far-field audio examples. This model had the same architecture but 40\% of the examples that the model was trained on were convolved with a randomly chosen room impulse response during training. We can see from Table \ref{tab:enhancer_results} that simple data augmentation can significantly improve performance on far-field audio without compromising the performance on near-field audio, implying that seq-to-seq models have a strong ability to learn from far-field examples.

Even with data augmentation, however, there is still a large gap between the WERs on near-field and far-field test sets. The bottom two rows of Table \ref{tab:enhancer_results} show the performance of the methods introduced in this paper on the same test sets. An $L^1$-distance penalty can lower the test set WER by 1.32\% absolute. Using a GAN enhancer can reduce the WER by an additional 1.07\%. Overall, the gap between near-field and far-field performance decreases by almost 27\% compared to the model that only uses data augmentation.

An additional benefit of our methods is that the $L^1$-distance penalty and GAN loss function act as regularizers which reduce generalization error on near field data. The enhancer models have lower WERs even on near-field data compared to the baseline models.

\section{CONCLUSION}
\label{sec:conclusion}

We introduced a GAN-based framework to train robust ASR models in a scalable, data-driven way, and showed that inducing invariance at the encoder embedding level considerably improves the recognition of simulated far-field speech by vanilla seq-to-seq models. This method has effectively imbued the seq-to-seq encoder with a far-field front-end. We anticipate that coupling the new framework with specialized trainable front-ends, such as WPE, would enhance robustness even more significantly.

\bibliographystyle{IEEEbib}
\bibliography{refs}

\end{document}